\documentclass[10pt,twocolumn,letterpaper]{article}

\usepackage[pagenumbers]{cvpr} 

\definecolor{cvprblue}{rgb}{0.21,0.49,0.74}
\usepackage[pagebackref,breaklinks,colorlinks,allcolors=cvprblue]{hyperref}
\usepackage{multirow}

\def\paperID{11337} 
\def\confName{CVPR}
\def\confYear{2025}

\title{Controllable Hand Grasp Generation for HOI and Efficient Evaluation Methods}

\author{
	Ishant,\quad Wu Rongliang,\quad Lim Joo Hwee,\\
	{\normalfont Institute for Infocomm Research (I$^{2}$R),}\\
	{\normalfont Agency for Science, Technology and Research (A*STAR),}\\
	{\normalfont 1 Fusionopolis Way, 21-01 Connexis (South Tower), 138632, Singapore.}\\
}

\begin{document}
\maketitle

\begin{abstract}

Controllable affordance Hand-Object Interaction (HOI) generation has become an increasingly important area of research in computer vision. In HOI generation, the hand grasp generation is a crucial step for effectively controlling the geometry of the hand. Current hand grasp generation methods rely on 3D information for both the hand and the object. In addition, these methods lack controllability concerning the hand's location and orientation. We treat the hand pose as the discrete graph structure and exploit the geometric priors. It is well established that higher order contextual dependency among the points improves the quality of the results in general. We propose a framework of higher order geometric representations (HOR's) inspired by spectral graph theory and vector algebra to improve the quality of generated hand poses. We demonstrate the effectiveness of our proposed HOR's in devising a controllable novel diffusion method (based on 2D information) for hand grasp generation that outperforms the state-of-the-art (SOTA). Overcoming the limitations of existing methods: like lacking of controllability and dependency on 3D information. Once we have the generated pose, it is very natural to evaluate them using a metric. Popular metrics like FID and MMD are biased and inefficient for evaluating the generated hand poses. Using our proposed HOR’s, we introduce an efficient and stable framework of evaluation metrics for grasp generation methods, addressing inefficiencies and biases in FID and MMD. 

\end{abstract}

\section{Introduction}

Capturing and understanding Hand-Object Interactions (HOI) has attracted increasing interest due to its wide range of applications in robotics, human-computer interaction, augmented reality/virtual reality, etc. 
Recent studies demonstrate that the synthesized HOI data with realistic affordance contact can serve as reliable resources for imitation learning in robotics, reducing the reliance on manually collected task-specific data and facilitating large-scale model learning~\cite{kapelyukh2023dall, yang2023learning, xue2024hoi}.

Controllable affordance HOI generation has become an increasingly important area of research in computer vision \cite{ye2023affordance, zhang2024hoidiffusion}. Generating affordance-based HOI data is a challenging task because it requires modeling physically plausible contact between hands and target objects with diverse geometric characteristics. Such geometric modeling is more effective in hand pose space, which can then be used as guidance for the overall HOI generation task.
For example, methods like \cite{zhang2024hoidiffusion, wang20243d} use the hand grasp generation methods \cite{taheri2020grab, christen2022d, zhu2021toward} as the guidance for overall HOI generation. However, the grasp generation methods typically synthesize HOI in a random manner and lack the ability to generate interactions at user-specified locations, leading to limited flexibility. Therefore, there is a need for a controllable hand pose generative method that can produce realistic hand grasps to improve the overall controllable HOI generation task.

Current hand grasp generation methods rely on the 3D information of object and the 3D hand grasp. Acquiring precise 3D information about an object's geometry is often challenging due to factors such as occlusions, complex surface textures, and limited sensor resolution. Therefore, in this work we target at generating controllable hand grasps with 2D data. We treat the hand grasp as the discrete graph structure and exploit the geometric priors \cite{chen2023gsdf, bronstein2017geometric} present in the hand grasp. It is well established that higher order contextual dependency among the points improve the quality of the results in general, \cite{matsune2024geometry, quan2021higher}. We propose a framework of higher order geometric representations (HOR's) for the hand grasp. Moreover, we represent the hand grasp as the spectrum of graph \cite{chung1997spectral}, intra-distance of points in graph and vector representations inspired by \cite{matsune2024geometry}.

We demonstrate the effectiveness of our proposed HOR's in devising a novel diffusion method (based on 2D information) for controllable hand grasp generation that outperforms the state-of-the-art (SOTA). Overcoming the limitations of existing methods: like lacking of controllability and dependency on 3D information. We build upon the same principal of \cite{ye2023affordance}, disentangling the HOI interaction into where and how, in which orientation and location is determined by an input visual prompt. Moreover, we further generalize our approach into a framework and open up the possibility of plug and play for various geometric representations \cite{lovasz2019graphs, sun2024mmd, southern2024curvature}.

At this point, it is crucial to ask for an appropriate metrics to evaluate generated hand grasps. The most popular choice for evaluating generative models is the FID (Fréchet Inception Distance) metric~\cite{heusel2017gans}. It is well known that FID is sensitive to the representation embedding space \cite{soloveitchik2021conditional}. For our case, our experiments demonstrate that FID is sensitive to the choice of color representation for hand grasps. Another choice for evaluating generated hand grasps is the MMD (Maximum Mean Discrepancy) for graph generation methods \cite{o2021evaluation}. However, MMD is unstable and limited by the expressivity of the choice of kernel \cite{southern2024curvature}. Our experiments show that precomputing and evaluating MMD is inefficient and biased. Using our proposed HOR’s, in this work, we introduce a set of efficient and stable metrics for evaluating the quality of the generated hand grasps.

In summary, the contributions of this work are fourfold. 
\begin{enumerate}
    \item A novel diffusion model for controllable hand grasp generation with just 2D data.
    \item A plug and play, framework for expressing various higher order geometric properties of hand pose for better generation of affordance hand grasps.
    \item A plug and play framework of efficient and unbiased evaluation metrics for hand grasp generation. To best of our knowledge, this is the first attempt to present hand grasp generation evaluation metrics. 
    \item A thorough quantitative and qualitative experimental comparison for the variety of components of proposed framework. 
\end{enumerate}

\section{Related Work}

\subsection{Hand Grasp Generation}

Researchers have advanced affordance hand pose generation. GrabNet \cite{taheri2020grab} enables 3D grasps for unseen objects, GanHand \cite{corona2020ganhand} predicts optimal grasps, and GraspTTA \cite{jiang2021hand} refines hand poses for consistent contact. D-grasp \cite{christen2022d} uses physics-based simulations, while Zhu et al.~\cite{zhu2021toward} focus on more natural, human-like poses. While these methods generate plausible hand poses for objects, they lack control over the placement of generated poses, limiting their applicability. Our proposed method addresses this by allowing user-specified locations for affordance hand-object interactions.

\subsection{Evaluation Metric for Hand Pose Generation}
Some recent studies have introduced metrics tailored to evaluate the realism of generated HOI \cite{tzionas2016capturing, hasson2019learning}. However, these metrics typically require 3D representations of both hands and the target object, making them unsuitable for direct application to 2D images and usually time-consuming to compute. The most popular choice for evaluating generative models is the FID (Fréchet Inception Distance) metric \cite{heusel2017gans} which is sensitive to the variation in input space \cite{soloveitchik2021conditional}.

Another line of work focuses on evaluation metrics for graph generative models, i.e., Maximum Mean Discrepancy (MMD) \cite{o2021evaluation}. MMD-based metrics have been critiqued for requiring numerous parameter and function choices and lacking stability guarantees \cite{o2021evaluation}. Our experiments show that MMD-based methods are inefficient and biased. In contrast, we propose efficient and unbiased hand pose generation metrics that are more stable.

\subsection{Diffusion Models}

Recent advances in diffusion methods have enhanced hand-object interaction generation.
 \cite{zhang2024hoidiffusion} introduced HOIDiffusion, a two-stage framework that synthesizes 3D hand-object geometry and uses a diffusion model to capture detailed interactions. \cite{ye2023affordance} developed an affordance-aware diffusion model to improve semantic consistency in interactions. While these methods focus on 3D HOI synthesis, few address hand pose generation. Our work emphasizes modeling geometric properties in diffusion models to generate affordance-controllable 2D hand poses without 3D geometry.

\begin{figure*}
  \includegraphics[width=1.0\textwidth]{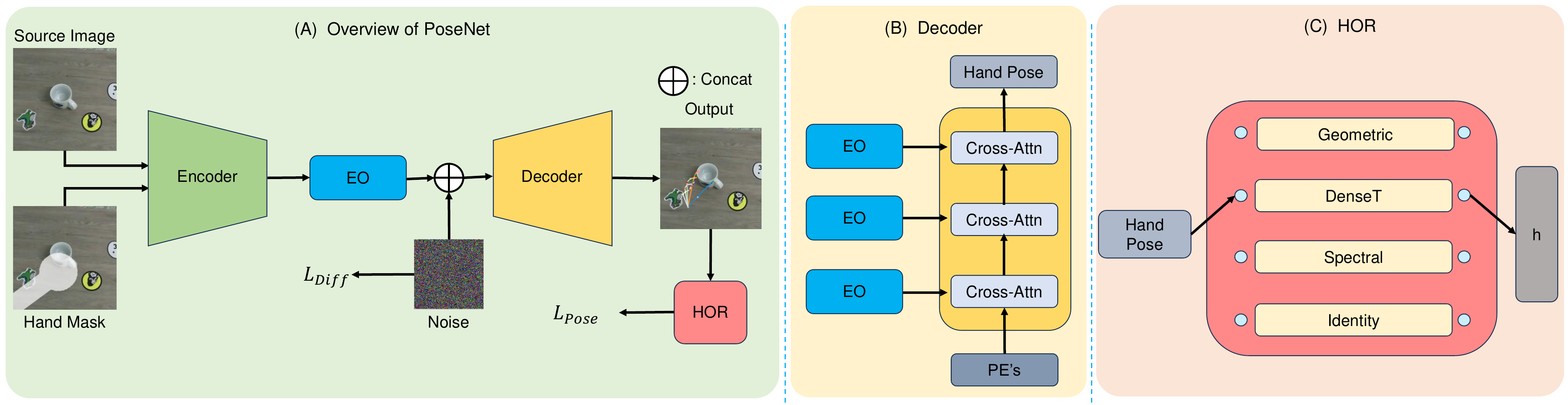}
  \centering
   \caption{Block (A), is the overview of our proposed controllable hand pose generation framework. In block (B), we present our position aware cross attention decoder. Block (C), is our proposed HOR's.}
  \label{fig:proposed1}
\end{figure*}

\section{Preliminary}

Diffusion models \cite{ho2020denoising} generate data by gradually adding Gaussian noise and then reversing the process, conditioned on an input context \( c \) that guides generation. The forward process adds noise as:

\begin{equation}
	q(\mathbf{x}_t \mid \mathbf{x}_{t-1}, c) = \mathcal{N}(\mathbf{x}_t; \sqrt{1 - \beta_t} \mathbf{x}_{t-1}, \beta_t \mathbf{I})
\end{equation}

where \( \beta_t \) controls the noise at each step \( t \). The reverse process, using a neural network \( \mathbf{\epsilon}_\theta(\mathbf{x}_t, t, c) \), denoises the data:

\begin{equation}
	p_\theta(\mathbf{x}_{t-1} \mid \mathbf{x}_t, c) = \mathcal{N}(\mathbf{x}_{t-1}; \mu_\theta(\mathbf{x}_t, t, c), \sigma_t^2 \mathbf{I})
\end{equation}

where \( \mu_\theta(\mathbf{x}_t, t, c) \) is the predicted mean.

The training objective for the diffusion model is to minimize the difference between the actual noise \( \mathbf{\epsilon} \) and the predicted noise \( \mathbf{\epsilon}_\theta(\mathbf{x}_t, t, c) \). This is achieved by optimizing the following loss function:
\begin{equation}
	L_{diff} = \mathbb{E}_{\mathbf{x}_0, \mathbf{\epsilon}, t, c} \left[ \left\| \mathbf{\epsilon} - \mathbf{\epsilon}_\theta(\mathbf{x}_t, t, c) \right\|^2 \right].
\end{equation}

This formulation ensures that the model learns to effectively denoise the data while taking the conditional input \( c \) into account, thus generating data that aligns with the given context.

\section{Method}


Hand poses have the geometric properties like finger joint angles, finger positions, palm orientation, hand shape, inter-finger relationships etc. Such geometric properties contains the higher order representation information of the hand joints and their orientations. It has been shown that using higher order representations helps in generating the better quality of pose in general \cite{matsune2024geometry, quan2021higher}. We show various strategies to compute higher order representations of hand pose. We take inspiration from graph theory, spectral graph theory \cite{chung1997spectral} and vector algebra. We show that using the higher-order representations improves the quality of generated hand poses. Later, we present a framework for designing generated hand pose evaluation metrics built upon our proposed descriptions of pose graph. We show that using the higher order representation is effective in evaluating the generated hand poses.

\subsection{Higher-Order Representations (HOR) for Hand Pose}
\label{sections:hor}
Let us consider the hand pose as the graph in which, bones as edges and bone joints and finger tips as nodes, call it pose graph. We denote $J$, as the set of all joints and finger tip points, where each point is a tuple, $(i,x,y)$, of index and $x,y$ coordinate. Without loss of generality (dimension of hand pose 2D, 3D, or different number of points), we consider the $21$ points hand pose model \cite{zimmermann2017learning}. We define the variety of higher order descriptor functions \cite{o2021evaluation} which take pose graph as input and produce descriptor vector that are used for expressing the hand poses. Different descriptor functions capture the higher order representations in a different manner. We identify each descriptor with a unique name for ease of reference as follows. We propose two novel HOR representation and adapt an existing descriptor for HOI generation and evaluation:

\subsubsection{Geometric} \citet{matsune2024geometry} proposed a geometric loss for body pose, which uses a combination of three descriptor functions. We adapt their loss terms to derive a descriptor function for hand pose, highlighting the adapted parameters. 

\textit{Root relative} We consider the wrist point as the root and compute the distance from the root to each point in $J$. The set of distances obtained is used as part of the total descriptor vector.

\textit{Bone Length} We consider all $14$ possible pairs of adjacent points (phalanges) in $J$ and compute the distance between them. This set of distances is appended to the total descriptor vector.

\textit{Phalanges Orientation} This term is derived from the Body Part Orientation term in \cite{matsune2024geometry}. In our case, it captures the orientation of each phalange by computing the cross product between two adjacent phalanges (vectors). Since we consider 2D hand pose, all such cross products or normals would be parallel to the image plane and redundant. Therefore, we consider the \( \sin(\theta) \) between two adjacent phalanges by computing the norm of the cross product. The set of \( \sin(\theta) \)'s is appended to the total descriptor vector.

\subsubsection{DenseT} Here we consider that the hand pose graph is a dense graph and derive a novel descriptor based on it. We compute the descriptor vector as the set of Euclidean distances between each pair of points in $J$. This descriptor vector is a superset of the sub-descriptor vectors obtained in \textit{Root relative} and \textit{Bone Length} of  \textbf{Geometric} descriptor. It can be considered a generalization of both sub-descriptors, where each point in $J$ is considered the root and each point is connected with a bone. We show in our experiments that it works well for the HOI generation.

\subsubsection{Spectral} We propose a novel descriptor for HOI generation based upon spectral properties of the hand pose graph. For a hand pose graph $G$, we define the weighted graph as described in \citet{chung1997spectral}:

\begin{equation}
	w(u, v) = \begin{cases} 
		-d(u, v) & \text{if } (u, v) \in G, \\
		d_v - d(u, v) & \text{if } u = v, \\
		0 & \text{otherwise}.
	\end{cases}
\end{equation}
Here, \( d(u, v) \) represents the Euclidean distance between points \( u \) and \( v \) ($\in J$), The term \( d_v \) represents the weighted degree of node \( v \), which is the sum of all distances \( d(u, v) \) for its neighboring nodes \( u \):

\begin{equation}
	d_v = \sum_{u \colon (u, v) \in G} d(u, v).
\end{equation}

The spectrum of the graph \( G \) is defined as the set of eigenvalues and eigenvectors of the Laplacian matrix \( \mathbf{L} ( = \mathbf{D} - \mathbf{W})\)  of the weighted adjacency matrix \( \mathbf{W} \). Where \( \mathbf{D} \) is the degree matrix, a diagonal matrix where each diagonal element \( d_{ii} \) represents the degree of node \( i \in J \). \textit{ For each hand pose, we represent its descriptor vector as the set of all its eigen vectors and eigen values}.

\subsubsection{Identity} We consider the identity descriptor vector that returns the coordinates of each point in $J$ as it is. It does not capture any higher order representations between hand pose points. We consider this transformation as the baseline of the descriptor functions.

\subsection{Controllable Hand Pose Generation}

We propose a diffusion model for generating controllable affordance 2D hand poses around an object  given in a input image. Our approach enables precise and context-aware 2D hand pose generation without relying upon any 3D geometric information of the object. In order to enhance the interaction between the hand and the object, we use the HOR of the hand pose. We use it as the reconstruction guidance for each diffusion step.

As shown in \cref{fig:proposed1}, our proposed architecture is a conditional encoder-decoder diffusion model designed for generating 2D hand poses at specified locations. We present two variations of the encoder: a U-Net encoder and a VAE encoder. Our proposed decoder is a transformer-based cross-attention decoder that effectively captures the relative positions of hand joints. We represent the specified location as the shape mask with $5$ parameters proposed by \cite{ye2023affordance}. We use the encoder to process the input image of an object and conditional location. Our decoder is a transformer-based cross attention module, which enhances the model's ability to focus on relevant features from the encoded input. The cross attention mechanism allows the decoder to attend to different parts of the encoded input dynamically, leading to more precise pose generation.

\subsubsection{Cross Attention Mechanism}
The cross attention mechanism in our decoder can be described using the following equations. Given an input query \( Q \), key \( K \), and value \( V \), the attention output is calculated as:

\begin{equation}
	\text{Attention}(Q, K, V) = \text{softmax} \left( \frac{QK^T}{\sqrt{d_k}} \right) V
\end{equation}

where \( d_k \) is the dimension of the key vectors. We use encoders output state to initialize the key and value of the attention layer. The cross attention mechanism allows the decoder to dynamically attend to different parts of the encoded input, facilitating more accurate hand pose generation. In order to capture the spatial relationships between joints, we apply learnable absolute positional encodings  corresponding to each point in $J$, \cite{dufter2022position} which allows the model to optimize the positional information during training.

\subsection{Loss Functions}

During the diffusion process, noise is incrementally added to the original (ground truth) hand pose, resulting in progressively noisier versions. To recover the original hand pose, we employ reverse dynamics, which denoise the hand pose at every step, moving from the noisy hand pose \(x_t\) back to the original hand pose \(x_0\). At each time step, the model predicts the noise component and subtracts it using the reverse dynamics to approximate the denoised hand pose. The reconstruction loss, which measures the difference between the denoised and ground truth hand poses, guides this process. By incorporating the similarity of HOR's in the reconstruction loss, we ensure that the model not only aligns individual joint positions accurately but also preserves the overall structural and functional relationships within the hand. This approach allows the model to produce hand poses that are both precise and realistic.

\subsubsection{Pose Reconstruction Loss:}
We denote a HOR (defined in \cref{sections:hor}) descriptor function as \(f\) which takes a hand pose as input and outputs a HOR descriptor vector. Following is the formula to compute the pose reconstruction loss for a given descriptor function $f$.

\begin{equation}
\mathcal{L}_{\text{pose}} = \mathbb{E}_{\mathbf{y}, \hat{\mathbf{y}}} \left[ \left\| f\mathbf{(y)} - \hat{f\mathbf{(y)}} \right\|^2 \right]
\end{equation}

where \( \mathbf{y} \) represents the ground truth hand pose, and \( \hat{\mathbf{y}} \) represents the predicted hand pose. We use the L$2$-norm operator to compute the distance between $\mathbf{y}$ and $\hat{\mathbf{y}}$.

\subsubsection{Combined Loss:}
To achieve robust hand pose estimation, we combine the diffusion loss and the pose reconstruction loss in our training objective:

\begin{equation}
	\mathcal{L} = \lambda_1 \mathcal{L}_{\text{diffusion}} + \lambda_2 \mathcal{L}_{\text{pose}}
\end{equation}

where \( \lambda_1, \lambda_2 \) are hyperparameter that balances the two loss terms. The diffusion loss \( \mathcal{L}_{\text{diffusion}} \) ensures the model effectively denoises the input, while the pose reconstruction loss \( \mathcal{L}_{\text{pose}} \) ensures the recovered hand pose has global spatial consistency.

\section{Evaluation Framework for Generative Hand Pose Models}
\label{section:posefid}
We propose a framework of the efficient evaluation metrics for hand pose generation methods. We use the HOR's of hand poses to derive the evaluation metrics.

For a descriptor function $f$, we represent the evaluation metric as $f$-FID. We transform each hand pose using the function $f$ and use the same principals as standard FID metric. We compute the mean  $\mu^f$  and covariance $\Sigma^f$ of the given population $P$ of hand poses as

\begin{equation}
	\mu^f = \frac{1}{|P|} \sum_{p \in P} f(p),
\end{equation}
\begin{equation}
	\Sigma^f = \frac{1}{|P|} \sum_{p \in P} (f(p) - \mu^f) (f(p) - \mu^f)^T
\end{equation}

For two populations of hand poses, with means  $\mu_1^f$ and $\mu_2^f$ and covariances  $\Sigma_1^f$ and $\Sigma_2^f$, the $f$-FID score is calculated as:

\begin{equation}
	f\text{-FID} = \| \mu_1^f - \mu_2^f \|^2 + \text{Tr}(\Sigma_1^f + \Sigma_2^f - 2(\Sigma_1^f \Sigma_2^f)^{1/2})
\end{equation}

We provide the technical details and pseudo code for our proposed evaluation metric in Appedix Sec. A.

\begin{figure*}
    \centering
  \includegraphics[width=1.0\textwidth]{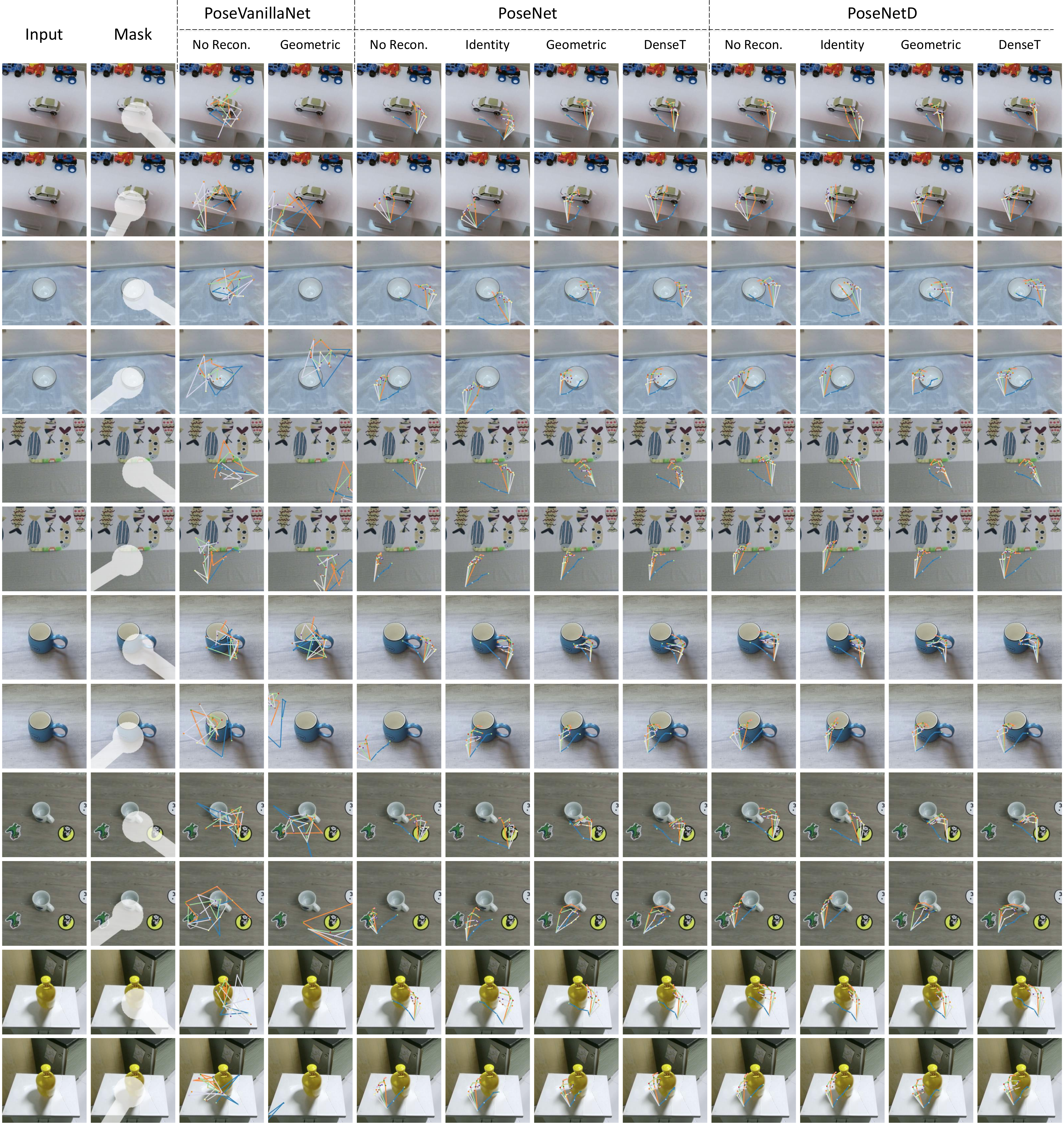}
   \caption{Qualitative comparison of our proposed method with the baseline method. The first column shows the Input images, and the second column shows the Mask images. The third column presents the results from the vanilla model (baseline). The fourth and fifth columns display the results from the PoseNet and PoseNetD models, respectively. Subsequent columns show the results of our proposed methods under different losses used while training. Each row corresponds to a different test image, demonstrating the performance of each method across a variety of input scenarios. Second row for each object shows the generated hand pose results from out of distribution orientation. We can see that our proposed method outperforms the vanilla method (columns 3,4). Using HOR based loss with proposed method outperforms the proposed method with Identity or no reconstruction loss, on qualitative results.}
   
  \label{fig:qualitative_res_all}
\end{figure*}

\begin{figure*}
	\centering
	\includegraphics[width=0.8\textwidth]{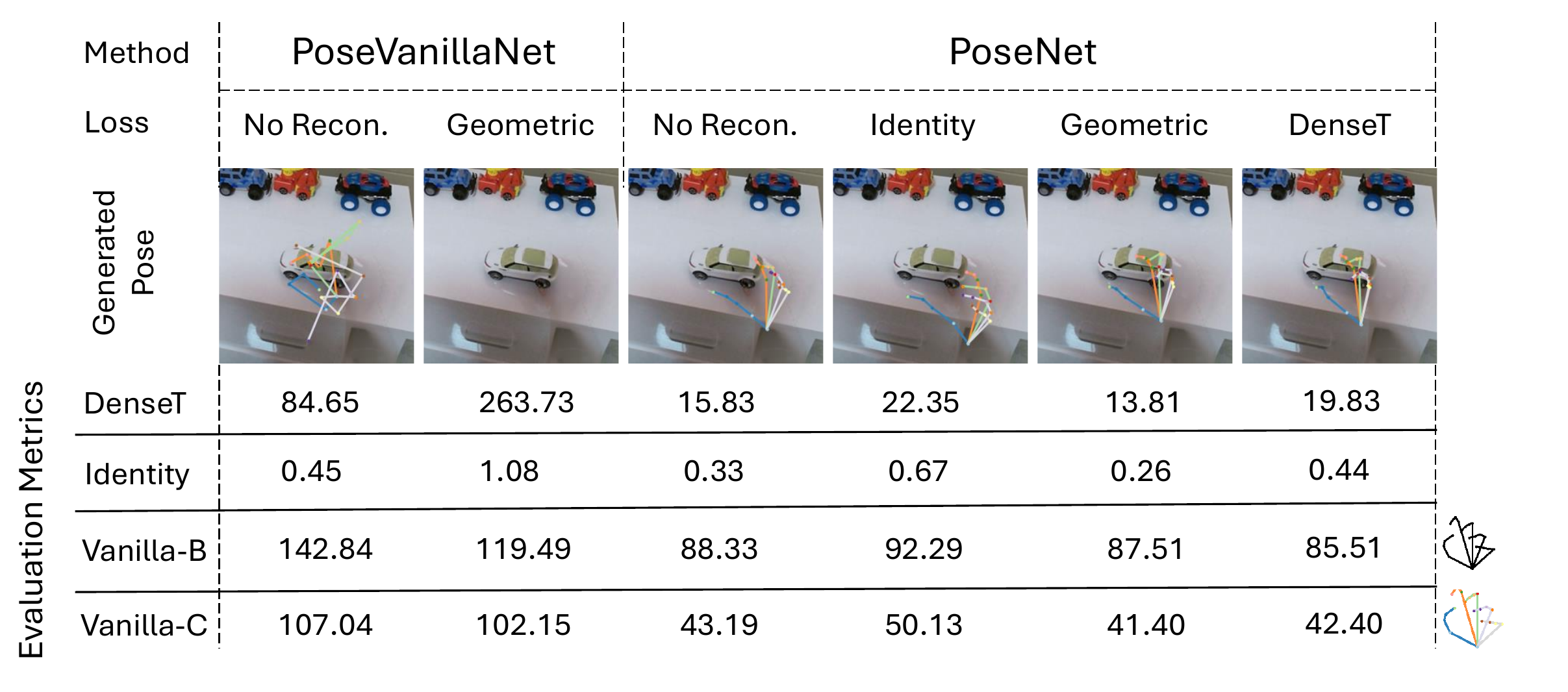}
	\caption{In this figure, we analyze the alignment between qualitative and quantitative results of various evaluation metrics. The numbers in each row correspond to evaluation metrics. Notably, row $2$ demonstrates that the DenseT metric aligns well with the generated poses, low value for good quality and high value for poor quality. Conversely, the Identity metric does not; the value in column $1$ $(0.45)$ is lower than that in column $4$ $(0.67)$, whereas it should be higher based on the qualitative alignment. Rows $4$ and $5$ present the quantitative results for FID computed on grayscale and colored images denoted as Vanilla-B and Vanilla-C. We can see that Vanilla-C aligns well with qualitative results then Vanilla-B, showing that FID is sensitive to rendering color choice, which impacts its stability.}
	
	\label{fig:eval_stable}
\end{figure*}

\renewcommand{\arraystretch}{1.08}
\begin{table}[ht]
\centering
\begin{tabular}{|l|c|c|c|}
\hline
\multirow{2}{*}{\textbf{Evaluation Metric}} & \multicolumn{3}{c|}{\textbf{Time (seconds)}} \\
\cline{2-4}
 & \multicolumn{2}{c|}{\textbf{Precompute}} & \textbf{Evaluation} \\
\cline{2-4}
 & \textbf{2k} & \textbf{20k} & \textbf{360} \\
\hline
\textbf{Identity-FID} & 0.6 & 5.2 & 0.005 \\
\textbf{Geometric-FID} & 3.1 & 25.1 & 0.08 \\
\textbf{Spectral-FID} & 4.1 & 33.4 & 1.57 \\
\textbf{DenseT-FID} & 9.7 & 85.0 & 1.77 \\
\textbf{MMD-Metric} & 32.83 & 2380.1 & 2.92 \\
\textbf{Vanilla-FID(GPU)} & 200.7 & 3307.4 & 13.24 \\
\textbf{Vanilla-FID(CPU)} & 683.0 & 6584.7 & 26.39 \\
\hline
\end{tabular}
\caption{Rows corresponds to the different evaluation metric types. Second column shows Pre-computation time for for 2k and 20k samples. Third column shows evaluation time for $360$ generated results. We can see that our proposed $f$-FID metrics (rows [1,4]) are much efficient than current state of the art evaluation metrics, i.e, FID and MMD.}
\label{table:eval_time}
\end{table}

\renewcommand{\arraystretch}{1.08}
\begin{table*}[ht]
\centering
\begin{tabular}{|l|c|c|c|c|c|c|c|}
\hline
\multirow{2}{*}{\textbf{Model}} & \multicolumn{7}{c|}{\textbf{Evaluation Metrics $\downarrow$ }} \\
\cline{2-8}
 & \multicolumn{4}{c|}{\textbf{$f$-FID}} & \multirow{2}{*}{\textbf{MMD}} & \multirow{2}{*}{\textbf{Vanilla-B}} & \multirow{2}{*}{\textbf{Vanilla-C}} \\
\cline{2-5}
 & \textbf{DenseT} & \textbf{Geometric} & \textbf{Spectral} & \textbf{Identity} & & & \\
\hline

\textbf{(No Recon. loss)} & & & & & & & \\
\cline{1-1}
\textbf{PoseNet} & 15.83 & 6.03 & 3.77 & 0.33 & 3.42 & 88.33 & 43.19 \\
\textbf{PoseNetD} & 22.15 & 6.67 & 4.77 & 0.60 & 3.57 & 92.71 & 53.53 \\
\textbf{PoseNetLdm} & 13.90 & 5.02 & 3.96 & \textbf{0.24} & 3.11 & 90.51 & 44.18 \\
\textbf{PoseNetLdmD} & 12.64 & 7.22 & 4.25 & 0.32 & 3.71 & 94.99 & 48.09 \\
\textbf{PoseVanillaNet} & 84.65 & 12.86 & 38.05 & 0.45 & 5.01 & 142.84 & 107.04 \\

\hline
\textbf{(Identity-loss)} & & & & & & & \\
\cline{1-1}
\textbf{PoseNet} & 22.35 & 7.33 & 4.64 & 0.67 & 3.74 & 92.29 & 50.13 \\
\textbf{PoseNetD} & 19.74 & 5.92 & 4.38 & 0.49 & 3.34 & 88.92 & 44.20 \\
\textbf{PoseNetLdm} & \textbf{12.24} & 7.05 & 4.49 & 0.36 & 3.66 & 106.12 & 56.90 \\
\textbf{PoseNetLdmD} & 19.54 & 6.45 & 4.77 & 0.44 & 3.49 & 100.03 & 56.45 \\
\textbf{PoseVanillaNet} & 66.56 & 11.45 & 30.58 & 0.43 & 4.67 & 139.77 & 103.17 \\
\hline
\textbf{(DenseT-loss)} & & & & & & & \\
\cline{1-1}
\textbf{PoseNet} & 19.83 & 6.75 & 4.36 & 0.44 & 3.58 & \textbf{85.42} & 42.40 \\
\textbf{PoseNetD} & 21.05 & 7.67 & 5.07 & 0.54 & 3.83 & 98.39 & 49.55 \\
\textbf{PoseNetLdm} & 16.16 & \textbf{4.80} & 4.79 & 0.31 & \textbf{2.98} & 97.65 & 52.36 \\
\textbf{PoseNetLdmD} & 15.91 & 6.78 & 4.47 & 0.40 & 3.59 & 88.68 & 47.04 \\
\textbf{PoseVanillaNet} & 162.10 & 23.73 & 70.13 & 1.01 & 6.80 & 97.60 & 72.99 \\
\hline
\textbf{(Geometric-loss)} & & & & & & & \\
\cline{1-1}
\textbf{PoseNet} & 13.81 & 5.25 & \textbf{3.61} & 0.26 & 3.19 & 87.51 & \textbf{41.40} \\
\textbf{PoseNetD} & 20.06 & 6.49 & 4.54 & 0.67 & 3.51 & 97.53 & 52.30 \\
\textbf{PoseNetLdm} & 14.88 & 6.80 & 4.60 & 0.34 & 3.59 & 97.09 & 53.12 \\
\textbf{PoseNetLdmD} & 14.12 & 6.33 & 4.83 & 0.37 & 3.46 & 92.64 & 50.81 \\
\textbf{PoseVanillaNet} & 263.73 & 24.92 & 81.46 & 1.08 & 6.97 & 119.49 & 102.15 \\
\hline
\end{tabular}
\caption{Evaluation Scores for different conditional hand pose generative models on All split of the HOI4D dataset. Columns under $f$-FID: DenseT, Geometric, Spectral, Identity represents evaluation metrics derived from our framework. Columns Vanilla-B, Vanilla-C are FID metric computed on grayscale, colored hand poses. The rows show different models and their variations, including PoseVanillaNet, PoseNet, PoseNetD, PoseNetLdm, and PoseNetLdmD, with configurations using identity, DenseT and Geometric loss functions. Rows are clustered with loss function type for training the models.}
\label{table:fid-scores-all}
\end{table*}


\section{Experiments}

We present a detailed quantitative evaluation and qualitative comparison of our proposed methods. All experiments are conducted on a machine with a single GPU (NVIDIA TITAN Xp), 12 GB of GDDR5X memory and a CPU (Intel(R) Xeon(R) CPU E5-1650 v4 @ 3.60GHz) with 128 GB RAM, running Ubuntu 20.04.1 and using the PyTorch framework. We utilize the HOI4D dataset \cite{Liu_2022_CVPR} due to its diverse objects and rich annotations. From this dataset, we consider a set of images featuring 'Grasp' and 'Carry' (contact) actions across all objects. We sample $20,000$ images, referred to as the 'All' dataset, is selected for our analysis.

We compare the results of different hand pose generation methods in our experiments using the following methods:  \textbf{PoseVanillaNet}: This diffusion model employs a 3-layer MLP as the decoder coupled with a U-net encoder. It serves as the baseline method for evaluating the performance of our proposed methods.
 \textbf{PoseNet}: Our proposed diffusion model that works in pixel space. It uses our proposed cross attention decoder with U-net encoder.
 \textbf{PoseNetLdm}: Our proposed latent diffusion model that works in latent space. Using same decoder as PoseNet with VAE encoder.
\textbf{PoseNetD, PoseNetLdmD}: PoseNet and PoseNetLdm respectively with deeper ($3$ layers) decoder network. We train all the models using the Adam optimizer with a learning rate of $1e-5$, weight decay of $0.01$.

\begin{figure}
    \centering
    \includegraphics[width=\columnwidth]{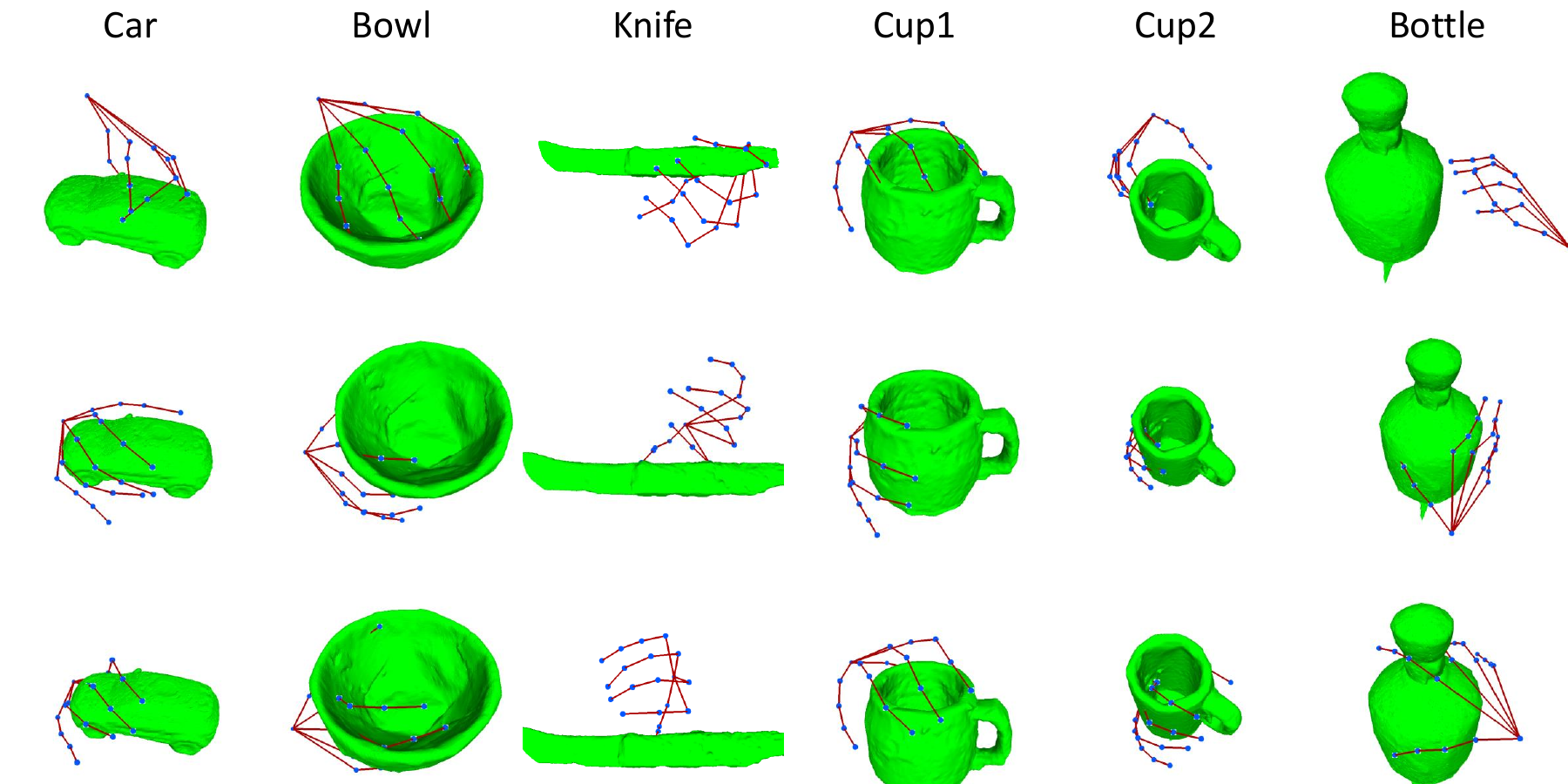}
    \caption{Grabnet hand grasp generation result for various objects in respective order from  \cref{fig:qualitative_res_all}.  }
    \label{fig:grabnet}
\end{figure}

In \cref{fig:qualitative_res_all}, we present qualitative results for various methods trained with different loss functions. We observe that using HOR's Geometric and DenseT significantly improves the contact consistency between the hand and the object. In contrast, using the Identity function or omitting the reconstruction loss results in relatively poorer quality. The second row for each object shows orientations that are out of the data distribution. Our proposed methods demonstrate good generalization to unseen orientations and locations. In  \cref{fig:grabnet}, we show the hand grasp results generated by GrabNet \cite{taheri2020grab} for the objects depicted in \cref{fig:qualitative_res_all}. GrabNet generates hand grasps at random locations, often leading to unrealistic outcomes. Additionally, it requires a 3D point cloud for each object to generate the hand grasps, whereas our method requires only a 2D RGB image of the object. We provide additional qualitative results in Appedix Sec. C. 

We use the HOR's: DenseT, Geometric, Spectral, and Identity to derive the evaluation metrics within our framework. We employ the MMD evaluation metric with Euclidean distance as the kernel \cite{o2021evaluation} along with the standard FID, using their official implementations.

In \cref{table:fid-scores-all}, we present the quantitative analysis of our proposed models (PoseNet, PoseNetLdm, PoseNetD, PoseNetLdmD) compared with the baseline PoseVanillaNet. We generated $360$ images using each method. The columns contain the evaluation metrics as described, and the rows list the proposed and baseline methods for hand pose generation. The rows are grouped based on the pose reconstruction loss. Our proposed models outperform the vanilla approach across all evaluation metrics. Additionally, it is evident that our HOR's representation-based loss performs better than the Identity representation for all evaluation metrics showing its efficacy to capture higher order geometric interactions.

\subsection{Efficiency of Evaluation Metrics}

We show the efficiency of our proposed evaluation metrics by comparing the pre-computation and evaluation times of various evaluation metrics in  \cref{table:eval_time}. We compute FID using both GPU and CPU: Vanilla-FID(GPU) and Vanilla-FID(CPU) and CPU versions for all other metrics. In  \cref{table:eval_time}, the Precompute column shows the time required for precomputing $2000$ (subset of 'All' dataset) and $20,000$ (All) samples, while the Evaluation column shows the evaluation time for $360$ generated samples, excluding model inference time. Our proposed $f$-FID metric evaluation framework achieves much better computation efficiency ($\approx 100$ times) in both precomputation and evaluation settings.

\subsection{Stability of Evaluation Metrics}

In \cref{fig:eval_stable}, we analyze the stability of the FID and Identity evaluation metrics. Vanilla-B and Vanilla-C are computed using the grayscale and colored hand poses shown in rows $4$ and $5$ in the bottom right of \cref{fig:eval_stable}. The quantitative results for Vanilla-C align better with the qualitative results (row $1$) than those for Vanilla-B, indicating that FID is sensitive to rendering color choice, which affects its stability. In row $3$ (Identity), the quantitative outputs of the Identity metric do not align well with the visually generated data (row $1$). In contrast, the evaluation metrics DenseT align well with the visually generated data. In \cref{table:fid-scores-all}, the evaluation metrics based on HORs: Geometric and Spectral follow the same trend as DenseT. Specifically, the values for PoseNetLDM and PoseVanillaNet (DenseT-loss) under the Vanilla-B column are $97.65$ and $97.60$, respectively, which contradict the qualitative results and imply that Vanilla-B is biased.

\subsection{Ablation Studies}

We refer to the  \cref{table:fid-scores-all} for studying the effectiveness of various component of our framework.

\subsubsection{Models}
We compare the different choices of encoder and decoder in our model. As we can see that our proposed model choices outperforms the Vanilla baseline for all possible losses in all the evaluation metrics.

\subsubsection{Pose Reconstruction Loss}
Each model is trained using various pose reconstruction loss functions derived from our proposed HOR's framework. As we can see that  using the pose reconstruction loss (Identity, DenseT, Geometric) is effective as compared to not using it (No Recon. loss). Moreover, We can observe that using the HOR representations: DenseT and  Geometric, outperforms Identity representation for pose reconstruction loss, showing the effectiveness of the HOR in capturing the geometric properties of hand pose.

\subsubsection{Evaluation Metric}
We compare various evaluation metrics using different HOR's: DenseT, Geometric, Spectral, and Identity. We notice that Identity representation is unstable. On the other hand, HOR's show more stability and aligns well with the visually generated data (as shown in \cref{fig:eval_stable}). Even among HOR based evaluation metric, Spectral shows better stability.

\section{Conclusion}

We introduce novel controllable hand grasp generation models based on 2D RGB input images of the target object. We demonstrate the effectiveness of the proposed HOR's representations in achieving high-quality results for hand grasp generation for out of distribution as well. Additionally, we show that state of the art evaluation metrics are unstable and inefficient for evaluating hand pose generation task. 
we introduce an efficient and stable evaluation metric, specifically designed for hand pose generation methods, based on our HOR framework.

{
    \small
    \bibliographystyle{ieeenat_fullname}
    \bibliography{sample}

\begin{thebibliography}{28}
\providecommand{\natexlab}[1]{#1}
\providecommand{\url}[1]{\texttt{#1}}
\expandafter\ifx\csname urlstyle\endcsname\relax
  \providecommand{\doi}[1]{doi: #1}\else
  \providecommand{\doi}{doi: \begingroup \urlstyle{rm}\Url}\fi

\bibitem[Bronstein et~al.(2017)Bronstein, Bruna, LeCun, Szlam, and
  Vandergheynst]{bronstein2017geometric}
Michael~M Bronstein, Joan Bruna, Yann LeCun, Arthur Szlam, and Pierre
  Vandergheynst.
\newblock Geometric deep learning: going beyond euclidean data.
\newblock \emph{IEEE Signal Processing Magazine}, 34\penalty0 (4):\penalty0
  18--42, 2017.

\bibitem[Chen et~al.(2023)Chen, Chen, Schmid, and Laptev]{chen2023gsdf}
Zerui Chen, Shizhe Chen, Cordelia Schmid, and Ivan Laptev.
\newblock gsdf: Geometry-driven signed distance functions for 3d hand-object
  reconstruction.
\newblock In \emph{Proceedings of the IEEE/CVF conference on computer vision
  and pattern recognition}, pages 12890--12900, 2023.

\bibitem[Christen et~al.(2022)Christen, Kocabas, Aksan, Hwangbo, Song, and
  Hilliges]{christen2022d}
Sammy Christen, Muhammed Kocabas, Emre Aksan, Jemin Hwangbo, Jie Song, and
  Otmar Hilliges.
\newblock D-grasp: Physically plausible dynamic grasp synthesis for hand-object
  interactions.
\newblock In \emph{Proceedings of the IEEE/CVF Conference on Computer Vision
  and Pattern Recognition}, pages 20577--20586, 2022.

\bibitem[Chung(1997)]{chung1997spectral}
Fan~RK Chung.
\newblock \emph{Spectral graph theory}.
\newblock American Mathematical Soc., 1997.

\bibitem[Corona et~al.(2020)Corona, Pumarola, Alenya, Moreno-Noguer, and
  Rogez]{corona2020ganhand}
Enric Corona, Albert Pumarola, Guillem Alenya, Francesc Moreno-Noguer, and
  Gr{\'e}gory Rogez.
\newblock Ganhand: Predicting human grasp affordances in multi-object scenes.
\newblock In \emph{Proceedings of the IEEE/CVF conference on computer vision
  and pattern recognition}, pages 5031--5041, 2020.

\bibitem[Dufter et~al.(2022)Dufter, Schmitt, and
  Sch{\"u}tze]{dufter2022position}
Philipp Dufter, Martin Schmitt, and Hinrich Sch{\"u}tze.
\newblock Position information in transformers: An overview.
\newblock \emph{Computational Linguistics}, 48\penalty0 (3):\penalty0 733--763,
  2022.

\bibitem[Hasson et~al.(2019)Hasson, Varol, Tzionas, Kalevatykh, Black, Laptev,
  and Schmid]{hasson2019learning}
Yana Hasson, Gul Varol, Dimitrios Tzionas, Igor Kalevatykh, Michael~J Black,
  Ivan Laptev, and Cordelia Schmid.
\newblock Learning joint reconstruction of hands and manipulated objects.
\newblock In \emph{Proceedings of the IEEE/CVF conference on computer vision
  and pattern recognition}, pages 11807--11816, 2019.

\bibitem[Heusel et~al.(2017)Heusel, Ramsauer, Unterthiner, Nessler, and
  Hochreiter]{heusel2017gans}
Martin Heusel, Hubert Ramsauer, Thomas Unterthiner, Bernhard Nessler, and Sepp
  Hochreiter.
\newblock Gans trained by a two time-scale update rule converge to a local nash
  equilibrium.
\newblock \emph{Advances in neural information processing systems}, 30, 2017.

\bibitem[Ho et~al.(2020)Ho, Jain, and Abbeel]{ho2020denoising}
Jonathan Ho, Ajay Jain, and Pieter Abbeel.
\newblock Denoising diffusion probabilistic models.
\newblock In \emph{Advances in Neural Information Processing Systems}, 2020.

\bibitem[Jiang et~al.(2021)Jiang, Liu, Wang, and Wang]{jiang2021hand}
Hanwen Jiang, Shaowei Liu, Jiashun Wang, and Xiaolong Wang.
\newblock Hand-object contact consistency reasoning for human grasps
  generation.
\newblock In \emph{Proceedings of the IEEE/CVF international conference on
  computer vision}, pages 11107--11116, 2021.

\bibitem[Kapelyukh et~al.(2023)Kapelyukh, Vosylius, and
  Johns]{kapelyukh2023dall}
Ivan Kapelyukh, Vitalis Vosylius, and Edward Johns.
\newblock Dall-e-bot: Introducing web-scale diffusion models to robotics.
\newblock \emph{IEEE Robotics and Automation Letters}, 8\penalty0 (7):\penalty0
  3956--3963, 2023.

\bibitem[Liu et~al.(2022)Liu, Liu, Jiang, Lyu, Wan, Shen, Liang, Fu, Wang, and
  Yi]{Liu_2022_CVPR}
Yunze Liu, Yun Liu, Che Jiang, Kangbo Lyu, Weikang Wan, Hao Shen, Boqiang
  Liang, Zhoujie Fu, He Wang, and Li Yi.
\newblock Hoi4d: A 4d egocentric dataset for category-level human-object
  interaction.
\newblock In \emph{Proceedings of the IEEE/CVF Conference on Computer Vision
  and Pattern Recognition (CVPR)}, pages 21013--21022, 2022.

\bibitem[Lov{\'a}sz(2019)]{lovasz2019graphs}
L{\'a}szl{\'o} Lov{\'a}sz.
\newblock \emph{Graphs and geometry}.
\newblock American Mathematical Soc., 2019.

\bibitem[Matsune et~al.(2024)Matsune, Hu, Li, Wen, Zhu, and
  Tan]{matsune2024geometry}
Ai Matsune, Shichen Hu, Guangquan Li, Sihan Wen, Xiantan Zhu, and Zhiming Tan.
\newblock A geometry loss combination for 3d human pose estimation.
\newblock In \emph{Proceedings of the IEEE/CVF Winter Conference on
  Applications of Computer Vision}, pages 3272--3281, 2024.

\bibitem[O'Bray et~al.(2021)O'Bray, Horn, Rieck, and
  Borgwardt]{o2021evaluation}
Leslie O'Bray, Max Horn, Bastian Rieck, and Karsten Borgwardt.
\newblock Evaluation metrics for graph generative models: Problems, pitfalls,
  and practical solutions.
\newblock \emph{arXiv preprint arXiv:2106.01098}, 2021.

\bibitem[Quan and Hamza(2021)]{quan2021higher}
Jianning Quan and A~Ben Hamza.
\newblock Higher-order implicit fairing networks for 3d human pose estimation.
\newblock \emph{arXiv preprint arXiv:2111.00950}, 2021.

\bibitem[Soloveitchik et~al.(2021)Soloveitchik, Diskin, Morin, and
  Wiesel]{soloveitchik2021conditional}
Michael Soloveitchik, Tzvi Diskin, Efrat Morin, and Ami Wiesel.
\newblock Conditional frechet inception distance.
\newblock \emph{arXiv preprint arXiv:2103.11521}, 2021.

\bibitem[Southern et~al.(2024)Southern, Wayland, Bronstein, and
  Rieck]{southern2024curvature}
Joshua Southern, Jeremy Wayland, Michael Bronstein, and Bastian Rieck.
\newblock Curvature filtrations for graph generative model evaluation.
\newblock \emph{Advances in Neural Information Processing Systems}, 36, 2024.

\bibitem[Sun and Fan(2024)]{sun2024mmd}
Yan Sun and Jicong Fan.
\newblock Mmd graph kernel: Effective metric learning for graphs via maximum
  mean discrepancy.
\newblock In \emph{The Twelfth International Conference on Learning
  Representations}, 2024.

\bibitem[Taheri et~al.(2020)Taheri, Ghorbani, Black, and
  Tzionas]{taheri2020grab}
Omid Taheri, Nima Ghorbani, Michael~J Black, and Dimitrios Tzionas.
\newblock Grab: A dataset of whole-body human grasping of objects.
\newblock In \emph{Computer Vision--ECCV 2020: 16th European Conference,
  Glasgow, UK, August 23--28, 2020, Proceedings, Part IV 16}, pages 581--600.
  Springer, 2020.

\bibitem[Tzionas et~al.(2016)Tzionas, Ballan, Srikantha, Aponte, Pollefeys, and
  Gall]{tzionas2016capturing}
Dimitrios Tzionas, Luca Ballan, Abhilash Srikantha, Pablo Aponte, Marc
  Pollefeys, and Juergen Gall.
\newblock Capturing hands in action using discriminative salient points and
  physics simulation.
\newblock \emph{International Journal of Computer Vision}, 118:\penalty0
  172--193, 2016.

\bibitem[Wang et~al.(2024)Wang, Xiang, Ding, and El~Saddik]{wang20243d}
Jiye Wang, Xuezhi Xiang, Shuai Ding, and Abdulmotaleb El~Saddik.
\newblock 3d hand pose estimation and reconstruction based on multi-feature
  fusion.
\newblock \emph{Journal of Visual Communication and Image Representation},
  101:\penalty0 104160, 2024.

\bibitem[Xue et~al.(2024)Xue, Luo, Chen, and Grauman]{xue2024hoi}
Zihui Xue, Mi Luo, Changan Chen, and Kristen Grauman.
\newblock Hoi-swap: Swapping objects in videos with hand-object interaction
  awareness.
\newblock \emph{arXiv preprint arXiv:2406.07754}, 2024.

\bibitem[Yang et~al.(2023)Yang, Du, Ghasemipour, Tompson, Schuurmans, and
  Abbeel]{yang2023learning}
Mengjiao Yang, Yilun Du, Kamyar Ghasemipour, Jonathan Tompson, Dale Schuurmans,
  and Pieter Abbeel.
\newblock Learning interactive real-world simulators.
\newblock \emph{arXiv preprint arXiv:2310.06114}, 2023.

\bibitem[Ye et~al.(2023)Ye, Li, Gupta, De~Mello, Birchfield, Song, Tulsiani,
  and Liu]{ye2023affordance}
Yufei Ye, Xueting Li, Abhinav Gupta, Shalini De~Mello, Stan Birchfield, Jiaming
  Song, Shubham Tulsiani, and Sifei Liu.
\newblock Affordance diffusion: Synthesizing hand-object interactions.
\newblock In \emph{Proceedings of the IEEE/CVF Conference on Computer Vision
  and Pattern Recognition}, pages 22479--22489, 2023.

\bibitem[Zhang et~al.(2024)Zhang, Fu, Ding, Liu, Tu, and
  Wang]{zhang2024hoidiffusion}
Mengqi Zhang, Yang Fu, Zheng Ding, Sifei Liu, Zhuowen Tu, and Xiaolong Wang.
\newblock Hoidiffusion: Generating realistic 3d hand-object interaction data.
\newblock In \emph{Proceedings of the IEEE/CVF Conference on Computer Vision
  and Pattern Recognition}, pages 8521--8531, 2024.

\bibitem[Zhu et~al.(2021)Zhu, Wu, Lin, and Sun]{zhu2021toward}
Tianqiang Zhu, Rina Wu, Xiangbo Lin, and Yi Sun.
\newblock Toward human-like grasp: Dexterous grasping via semantic
  representation of object-hand.
\newblock In \emph{Proceedings of the IEEE/CVF International Conference on
  Computer Vision}, pages 15741--15751, 2021.

\bibitem[Zimmermann and Brox(2017)]{zimmermann2017learning}
Christian Zimmermann and Thomas Brox.
\newblock Learning to estimate 3d hand pose from single rgb images.
\newblock In \emph{Proceedings of the IEEE international conference on computer
  vision}, pages 4903--4911, 2017.

\end{thebibliography}
}


\end{document}